\documentclass[11pt]{article}
\ifdefined\XeTeXversion\else\pdfoutput=1\fi
\usepackage[T1]{fontenc}
\usepackage{lmodern}
\usepackage{microtype}
\usepackage{amsmath,amssymb,amsthm}
\usepackage{booktabs}
\usepackage{graphicx}
\usepackage[letterpaper,margin=1.05in]{geometry}
\usepackage[colorlinks=true,linkcolor=blue,citecolor=blue,urlcolor=blue]{hyperref}
\usepackage{url}
\theoremstyle{definition}
\newtheorem{problem}{Problem}
\newtheorem{definitionx}{Definition}

\title{Can Event Detectors Leave the Broadcast Camera?\\ Evidence from Tennis}
\author{David Soeiro-Vuong \qquad Andr\'e-Louis Rochet\\[2pt] Racine.ai, Paris, France}
\date{October 2026}

\begin{document}
\maketitle

\begin{abstract}
Ball bounces and racket hits are the events from which tennis statistics are derived, and on
monocular video the two are easy to confuse. We release a benchmark of 345 bounce and hit
annotations at frame level on eight clips that mix broadcast, fixed elevated, phone and court-level viewpoints, with a frozen development and test split, a confusion-aware matcher, clip-level bootstrap
intervals and a blind second annotation of three clips. On it we score supervised precise spotters
with published broadcast weights, two zero-shot vision-language models, and RallyArb, an arbitration
layer over a ball track and player poses that fits no parameter on event labels. The supervised
spotters score near-perfectly on broadcast footage and almost nothing on phone and court-level footage, where the
pooled bounce F1 of E2E-Spot is 0.130. RallyArb keeps a bounce F1 above 0.8 on the three phone and court-level clips, two of which were used to tune it, and finds 8 of 11 bounces on the held-out one, but it places
hits too far from the contact frame to match human timing. The vision-language models detect events
on every viewpoint and miss the contact frame by about 0.1~s. Fine-tuned leave-one-clip-out from its
broadcast weights, the spotter reaches about 0.8 bounce F1 on two phone and court-level clips and stays at 0.2
on a grazing phone angle. The annotations, the evaluator and the frozen predictions of every scored
row are released.
\end{abstract}

\section{Introduction}
Automatic tennis analytics was built for the broadcast setting, with multi-camera line calling
(Owens et al., 2003), court registration from broadcast angles (Farin et al., 2004) and ball trackers
trained on professional footage (Huang et al., 2019; Tarashima et al., 2023). Amateur players who
film themselves with one phone on a tripod need the same analytics from a viewpoint that broadcast systems never see. For them the useful output is the event stream derived from the ball
track: where the ball bounced, when it was struck, and how the two alternate within a rally.

Bounces and hits look alike on a monocular track. Both are sharp discontinuities of the image
trajectory, and a detector tuned for one fires on the other. A hit emitted as a bounce draws a ball
mark where no ball landed. We therefore score the two classes jointly and count a prediction of the
wrong class as an error on both.

We found no public corpus that measures whether event detectors transfer from broadcast footage to
the viewpoints amateurs use. Precise-spotting benchmarks in tennis are built from broadcasts (Hong et
al., 2022; Liu et al., 2025), and racket-sport benchmarks target ball and racket tracking (Dong et
al., 2026). We built a diagnostic benchmark for this question, designed to measure transfer across
viewpoints.

The paper makes four contributions. First, we release a benchmark of eight clips and 345 frame-level
bounce and hit events across clay, grass and hard courts at 25 to 60 frames per second, with a
versioned annotation policy, exclusion tags, a frozen split, a blind second annotation of three clips
and an evaluator that needs only the Python standard library (Section~\ref{sec:banc}). Second, we
score supervised spotters with published weights, two vision-language models and a training-free
arbitration layer under one fixed protocol, with one test execution per row
(Sections~\ref{sec:familles} and~\ref{sec:protocole}). Third, we observe, on two or three clips per
camera type, that the ranking of these families changes with the camera viewpoint and the matching
tolerance, and we separate the clips each row could have been tuned on from those it could not
(Section~\ref{sec:resultats}). Fourth, a leave-one-clip-out fine-tuning measures what the 345 events
teach a supervised spotter.

\section{Related work}
\paragraph{Ball tracking in racket sports.} TrackNet introduced heatmap-based tracking of small fast
balls in broadcast tennis (Huang et al., 2019), and later versions extended it to shuttlecocks and
added trajectory rectification and motion attention (Sun et al., 2020; Chen and Wang, 2023; Raj et
al., 2024). WASB is a strong baseline for ball detection and tracking across sports (Tarashima et al.,
2023). The arbitration layer evaluated here consumes such a tracker as an interchangeable front end,
in our pipeline a Kalman-filtered (Kalman, 1960) single-class detector.

\paragraph{Event and stroke annotation in racket sports.} TenniSet annotates dense fine-grained
events, including hits and bounces, in broadcast tennis (Faulkner and Dick, 2017). TTNet learns
table-tennis bounces and net hits from densely annotated video (Voeikov et al., 2020), and P2ANet
annotates dense table-tennis actions in broadcasts (Bian et al., 2022). MonoTrack segments badminton
rallies from hit events detected with trajectory and pose cues (Liu and Wang, 2022), and ShuttleSet
labels badminton strokes at scale (Wang et al., 2023). EventAnchor reduces the human effort of
annotating racket-sport events (Deng et al., 2021). Each of these resources is drawn from broadcast
or fixed match footage of its own sport.

\paragraph{Precise event spotting.} SoccerNet-v2 popularized action spotting (Deli\`ege et al., 2021),
and its ball action spotting track tightened the target to ball state changes (Cioppa et al., 2023,
2024; Giancola et al., 2025). E2E-Spot and T-DEED learn frame-level spotters end to end (Hong et al.,
2022; Xarles et al., 2024). The tennis dataset of E2E-Spot extends Vid2Player (Zhang et al., 2021) to
28 US Open and Wimbledon broadcast matches with frame-accurate bounce, swing and serve contacts on the
near and far court (Hong et al., 2022). F\textsuperscript{3}Set benchmarks fast and fine-grained
tennis events (Liu et al., 2025), few-shot variants reduce the need for labels (Liu et al., 2026), and
a survey covers the field (Xu et al., 2025). RacketVision benchmarks ball and racket analysis across
racket sports without bounce and hit events (Dong et al., 2026).

\paragraph{Temporal localization by vision-language models.} Video language models answer questions
about when an event happens, but their temporal predictions are inconsistent under small changes of
the query (Jung et al., 2025). We test whether a frame-indexed prompt is enough to place a ball
contact within a few frames.

\paragraph{Court registration and structured decoding.} Court-model calibration is mature for
broadcast angles (Farin et al., 2004), and line detection on amateur recordings has been studied where
lines are occluded or seen from low angles (Agrawal et al., 2024). The arbitration layer needs no
court registration. It reads trajectory directions with Theil--Sen slopes (Theil, 1950; Sen, 1968)
and decodes labels with a Viterbi-style dynamic program (Viterbi, 1967).

\section{Problem statement}
A clip has $N$ frames at rate $\nu$ frames per second. Its ground truth is a set of events
$G=\{(g_j,y_j)\}$ with $y_j\in\{\mathrm{B},\mathrm{H}\}$, where B is a floor bounce, H a racket contact
and $g_j$ the first frame of contact. Each event may carry a tag from a fixed vocabulary (serve, toss,
dead, far, offscreen, replay and others), and the scoring view decides which tags are excluded.

\begin{problem}[Joint event labeling]
Given a monocular clip, emit a set of predictions $P=\{(f_i,\ell_i)\}$ with
$\ell_i\in\{\mathrm{B},\mathrm{H}\}$ such that every untagged ground-truth event is matched by a
prediction of its class within a tolerance of $\tau=\mathrm{round}(0.15\,\nu)$ frames.
\end{problem}

\begin{definitionx}[Confusion-aware F1]
One greedy one-to-one matcher pairs predictions with ground-truth events of both classes at once,
nearest frame first, within $\tau$. A matched pair with $\ell_i\neq y_j$ is a confusion: it counts as
a false positive of class $\ell_i$ and a false negative of class $y_j$. An unmatched prediction is
spurious. Excluded ground-truth events are removed after matching, so a prediction matched to one of
them is ignored. For class B,
\[
P_\mathrm{B}=\frac{\mathrm{TP}_\mathrm{B}}{\mathrm{TP}_\mathrm{B}+\mathrm{conf}_{\mathrm{H}\to\mathrm{B}}+\mathrm{spur}_\mathrm{B}},\qquad
R_\mathrm{B}=\frac{\mathrm{TP}_\mathrm{B}}{\mathrm{TP}_\mathrm{B}+\mathrm{FN}_\mathrm{B}+\mathrm{conf}_{\mathrm{B}\to\mathrm{H}}},
\]
and F1 is their harmonic mean; class H is symmetric. Multi-clip scores are micro-averages of the
counters over clips.
\end{definitionx}

The definition penalizes a confusion twice, once in each class. A system that never confuses the
classes but misses events loses recall only, and a system that labels discontinuities with the wrong
class loses precision and recall together. The tolerance of 0.15~s is wider than the 1 to 2 frames
used by precise-spotting benchmarks (Hong et al., 2022); Result~6 reports every row at tolerances from
0 to 0.5~s.

\section{The benchmark}\label{sec:banc}
\begin{figure}[!htbp]
\centering
\includegraphics[width=\linewidth]{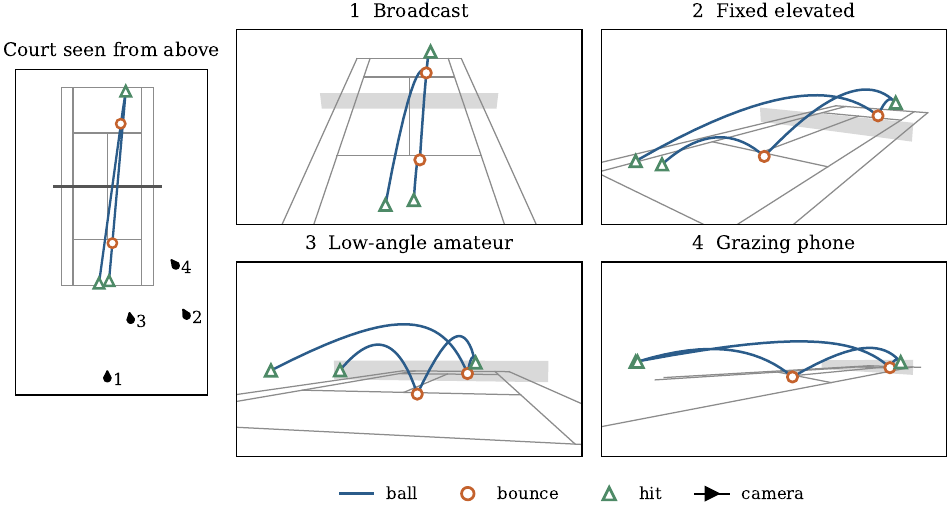}
\caption{The four viewpoint types of the benchmark, illustrated with one simulated rally segment: two
exchanges with ballistic flights on a court of regulation size, projected through four virtual pinhole
cameras one per viewpoint type (left, numbered). From broadcast height every bounce and hit occupies
its own region of the image. From an elevated side camera the far court is foreshortened. From the two low cameras the far bounce and the far hit fall within a few pixels of each other near the horizon, while
the near events fill the frame. The figure is synthetic and contains no benchmark data; the same
viewpoints on real frames are in Figure~\ref{fig:reel}. Each view is framed on the rally.}
\label{fig:schema}
\end{figure}

\begin{figure}[!htbp]
\centering
\includegraphics[width=\linewidth]{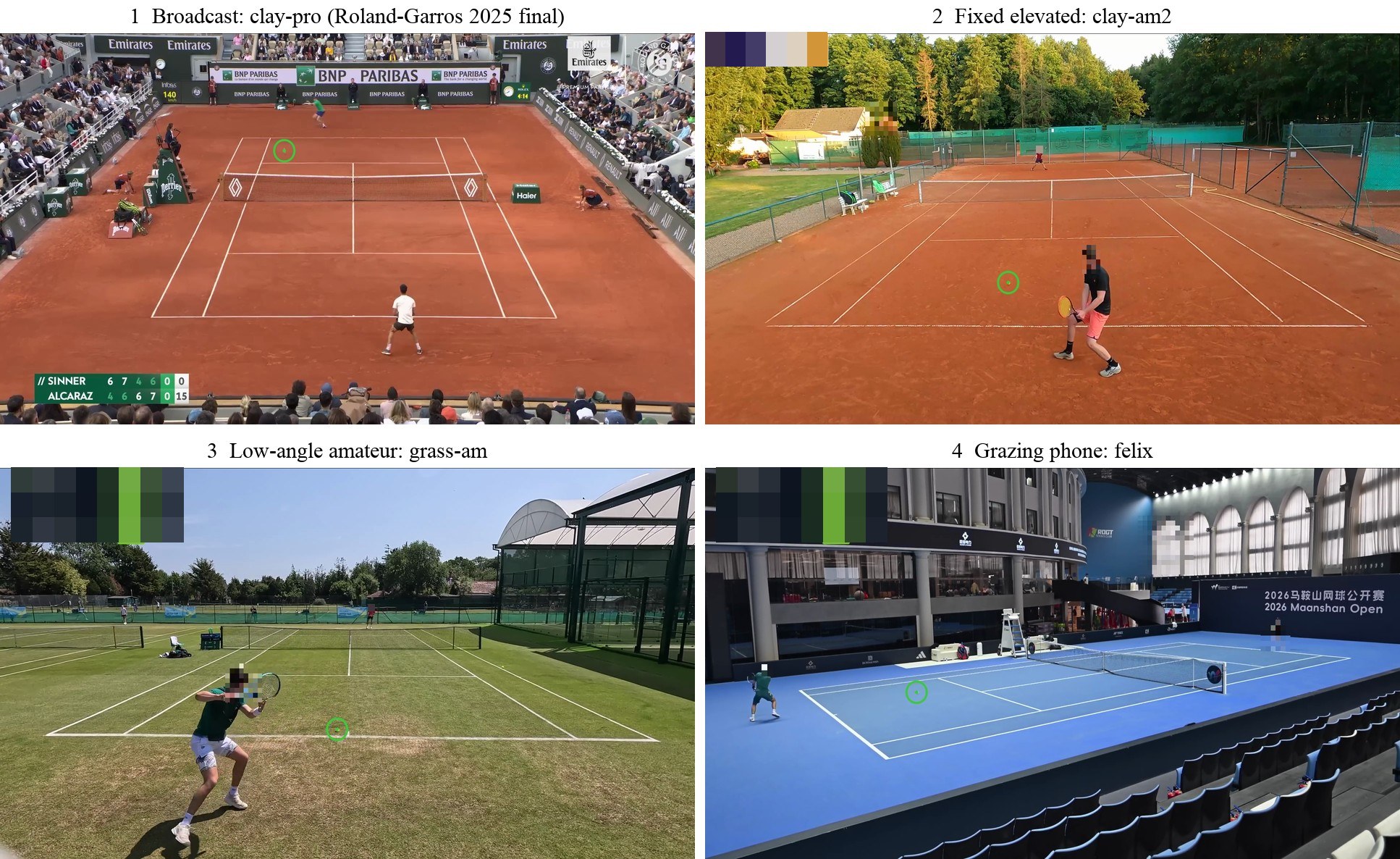}
\caption{The four viewpoint types of Figure~\ref{fig:schema} on real benchmark frames, each at a
ground-truth bounce (ball circled): 1 the 2025 Roland-Garros final at 25 frames per second, 2 an
elevated fixed camera on an outdoor clay court, 3 a low-angle phone on grass, bounce at the centre of
the court, 4 a static phone at a grazing angle, bounce on the near half.  Sources: the official Roland-Garros highlights (1) and videos published on YouTube by the players (2--4); faces and on-screen names are pixelated in frames 2--4.}
\label{fig:reel}
\end{figure}

\paragraph{Corpus.} Table~\ref{tab:corpus} lists the eight clips, Figure~\ref{fig:schema} illustrates
the four viewpoint types and Figure~\ref{fig:reel} shows them on real benchmark frames. They cover
three surfaces, frame rates from 25 to 60 frames per second, professional broadcast, court-level
practice, and amateur fixed, elevated, low-angle and grazing viewpoints. Five clips are cuts of about
40~s from longer recordings. The grass broadcast compilation is cut at 22~s, after its first rally and
before a slow-motion replay. Demo3 is a 13~s rally window of a longer broadcast, and felix the first
31~s of a phone recording.

\begin{table}[!htbp]
\centering\small\setlength{\tabcolsep}{4pt}
\resizebox{\linewidth}{!}{%
\begin{tabular}{lllrrrrl}
\toprule
Clip & Split & Camera and setting & fps & B & H & Events & Tags present\\
\midrule
demo3 & dev & broadcast oblique, pro rally & 50 & 10 & 9 & 19 & serve, toss\\
felix & dev & static phone, grazing angle, amateur & 60 & 18 & 22 & 40 & dead, serve\\
clay-am & dev & fixed camera, indoor clay, amateur & 60 & 22 & 18 & 40 & dead, serve\\
hard-pro & dev & court-level practice, pro (near-only) & 60 & 27 & 34 & 61 & dead, far, offscreen, toss\\
clay-pro & test & broadcast, clay, pro & 25 & 27 & 25 & 52 & dead, serve\\
clay-am2 & test & elevated fixed camera, outdoor clay, amateur & 59.94 & 25 & 27 & 52 & serve, toss\\
grass-pro & test & broadcast compilation, grass, pro & 60 & 12 & 14 & 26 & dead, serve\\
grass-am & test & low-angle amateur, grass (near-only) & 60 & 24 & 31 & 55 & dead, far, serve, toss\\
\midrule
Total & & & & 165 & 180 & 345 & \\
\bottomrule
\end{tabular}}
\caption{The benchmark corpus. Counts are those of the released ground-truth sidecars, tagged events
included. For the per-camera reading the clips are grouped as broadcast (demo3, clay-pro, grass-pro),
fixed elevated (clay-am, clay-am2) and phone or court level (felix, hard-pro, grass-am). Generated by
\texttt{make\_figures.py} from \texttt{data/gt/}.}
\label{tab:corpus}
\end{table}

\paragraph{Annotation policy.} Every physical contact is annotated at its first frame of contact,
including events outside the rally. Those carry a tag, such as serve, toss, dead, offscreen, far or
replay, and scoring excludes them by tag. Detecting a real serve is therefore neither rewarded nor
punished, and no real event is dropped from the files. The policy is versioned with the ground truth.
Version 2 widened the toss tag to every pre-serve ritual after an audit of spurious predictions on the
development clips, and the same audit filled annotation gaps it revealed and retimed three
development events by 8 to 15 frames. Hard-pro and grass-am are scored on the near half only, because
the far half of the court cannot be verified on such footage, even by a human annotator. The first
author annotated all eight clips.

\paragraph{Inter-annotator agreement.} A second annotator, a colleague at Racine.ai with no access to
the first annotation or to any prediction, re-annotated clay-am, grass-am and clay-pro from the raw
clips, the frozen policy and a written protocol, and delivered the three files before any discussion.
Scored with the benchmark matcher at the paper tolerance, the rally-only view agrees at 1.000 bounce
F1 and 0.957 hit F1 over 97 events. Labels agree on all 130 matched pairs, with no bounce and hit
confusion, and the two annotators agree on the player and the stroke of every matched hit. Over all
130 matched pairs, frames differ by 0.24 on average and by 3 at most, and 79\% coincide exactly; in
the rally-only view the largest difference is 2 frames. Over all annotated events, tags included,
agreement is 0.935 and 0.922; the remaining disagreements are far-half events on grass-am and
consecutive dead bounces on clay-am. The two hits on which the annotators disagree beyond the
tolerance, at 11 and 15 frames, are two of the three events the first annotator retimed during the
development audit, and the second annotator places both at their original frames. The second
annotator reports that part of the serve and dead tags were applied by a script from two written
rules on hand-entered events. The released ground truth was not modified.

\paragraph{Matching, pooling and uncertainty.} Scoring follows Definition~1. Clips are the resampling
unit of a bootstrap with 10\,000 iterations and a fixed seed, because events within a clip are
correlated. With four clips per split the 95\% intervals are wide. They are reported to prevent
over-reading pooled differences and do not support claims of significance.

\paragraph{Frozen split.} The development split holds demo3, felix, clay-am and hard-pro; tuning is
allowed on it. The test split holds clay-pro, clay-am2, grass-pro and grass-am and is scored once per
system version. Its events were annotated on 2026-09-03, before any tuning began, and its pipeline
outputs were not inspected during development. On 2026-09-05 a tag-only pass added eight toss events
and retagged two, without moving any rally event.

\paragraph{Replay fixtures.} The source videos are not redistributed. For each clip the release
includes a replay fixture with the per-frame tracked ball centre, a validity mask and two
identity-locked player poses with 17 keypoints. A method that works downstream of detection and
tracking can be evaluated from the fixtures alone; a method that reads pixels needs the videos.

\section{Detector families}\label{sec:familles}
\paragraph{Supervised precise spotters.} E2E-Spot (Hong et al., 2022) is run with its published
tennis weights (RegNetY-008) and T-DEED (Xarles et al., 2024) with the small tennis checkpoint
released in its code repository, both without retraining. Their six classes (near and far court
bounce, swing and serve) are mapped to B for the bounce classes and H for swing and serve. Inputs
follow the published recipe: frames resized to $398\times224$ at 25 to 30 frames per second (one frame
in two for our 50 and 60 frames per second clips) and a centre crop of 224 pixels. Of the two released
T-DEED checkpoints, the small one, designated beforehand, is reported. The softmax thresholds, 0.5
for E2E-Spot and 0.3 for T-DEED, were fixed before any result was read, with non-maximum suppression
over $0.15\,\nu$ frames.

\paragraph{Vision-language models.} Gemini~3.8 Flash (proprietary) and Qwen3-VL-235B-A22B (open
weights) receive windows of consecutive frames at about 12 frames per second and 640 pixels wide, with
the absolute frame index burned into each image. Windows hold 50 frames for Gemini and 32 for Qwen,
the latter a provider limit. One fixed prompt gives the annotation policy's definition of each event,
at temperature 0. The model and the sampling rate were chosen on development data. Two protocol
defects found there, a truncated answer and a brittle JSON reader, were fixed before the test pass,
and the prompt was not edited after a result was read.

\paragraph{RallyArb.} The arbitration layer runs downstream of a ball tracker and a pose estimator,
both trained models used unchanged, and fits no parameter on event labels. Five generators propose
unlabeled candidates: a curve-fit bounce detector, a wrist-speed hit detector, local extrema of the
image height, sharp turns of the velocity vector, and a near-court vertical-velocity knee. Candidates
within 0.18~s form one cluster. Each cluster gathers trajectory evidence from robust slopes on real
detections only, and a measurement that cannot be read casts no vote. Contact evidence is the distance
from the independently tracked ball to the nearest confident wrist, in units of that player's box
height. A rule set removes the bounce hypothesis from every cluster with hit evidence and the hit
hypothesis from every cluster with clear floor evidence, before any score is computed. A dynamic
program then places the remaining labels under the constraint that a rally alternates hits and bounces
at a beat estimated per clip. Every constant is a ratio of the frame rate, the frame size or the
player-box height. The constants were fixed on demo3 and then refined on the four development clips.
Appendix~\ref{app:arb} gives the rules and constants.

\paragraph{Fine-tuned spotter.} To ask whether the corpus teaches the missing viewpoints, E2E-Spot
with its smaller published weights (RegNetY-002, Radosavovic et al., 2020) is fine-tuned
leave-one-clip-out on the released events of the seven other clips. The published head is kept and
its six outputs are summed into bounce and hit probabilities, so that epoch zero reproduces the
published model exactly. Training uses AdamW (Loshchilov and Hutter, 2019) at a learning rate of
$10^{-4}$ for ten epochs of 256 clips, keeps the last epoch and thresholds at 0.5. All settings were
fixed before the first fold and each fold ran once. The resulting models measure learnability on this
corpus; none of them is a deployable detector.

\section{Evaluation protocol}\label{sec:protocole}
All rows are scored by the same evaluator with tolerance $\mathrm{round}(0.15\,\nu)$ and the rally-only
view, which excludes the tags serve, toss, far, dead, offscreen and replay. The test split carries 72
of its 140 rally-only events on two broadcast clips. Any number pooled over the test split therefore
favours methods that are strong on broadcast footage, and the per-camera grouping of
Table~\ref{tab:camera} is the primary reading. The rally-only view excludes dead bounces, the repeated
bounces after a point ends, which RallyArb's alternation decoder cannot emit by design. Scoring them
as well changes the pooled test bounce F1 of RallyArb from 0.621 to 0.634 and that of E2E-Spot from
0.776 to 0.750, so the exclusion does not favour RallyArb.

The order of events fixes what each row could have seen, and Table~\ref{tab:info} lists it per row.
The test ground truth was frozen on 2026-09-03. RallyArb was refined on the development split from
2026-09-05 and scored on test once on 2026-09-07. Before that run, the annotator applied the version-2
tag pass to the test ground truth without seeing any prediction, which added eight toss events and
retagged two; one test sidecar keeps its version-1 header. The external rows were run on 2026-10-15 to
2026-10-17 under a protocol written on 2026-10-15, at the time of their single test pass. The
leave-one-clip-out fine-tuning was specified on 2026-10-15 and run afterwards under settings fixed
before its first fold.

\begin{table}[!htbp]
\centering\small
\begin{tabular}{p{3.1cm}p{3.6cm}p{5.2cm}c}
\toprule
Row & Labels from this corpus & Choices made on development clips & Test runs\\
\midrule
RallyArb & development, through its refinements & constants refined on all four & 1\\
E2E-Spot, T-DEED & none & none (thresholds fixed a priori) & 1\\
Gemini 3.8 Flash & none & model choice, sampling rate, two protocol fixes & 1\\
Qwen3-VL-235B & none & sampling rate inherited from Gemini & 1\\
E2E-Spot fine-tuned & the seven other clips, test clips included & none & 1 per fold\\
\bottomrule
\end{tabular}
\caption{Information available to each row. A development clip is out of sample for every row except
RallyArb and Gemini. The fine-tuned spotter's test folds train on the labels of the three other test
clips. Diagnostic test executions not used for any choice: the large T-DEED checkpoint, the zero-shot
RegNetY-002 spotter, and a full-frame spotter run.}
\label{tab:info}
\end{table}

\section{Results}\label{sec:resultats}

\begin{table}[!htbp]
\centering\small
\begin{tabular}{lcccccc}
\toprule
& \multicolumn{2}{c}{Broadcast (3 clips)} & \multicolumn{2}{c}{Fixed elevated (2)} & \multicolumn{2}{c}{Phone or court level (3)}\\
\cmidrule(lr){2-3}\cmidrule(lr){4-5}\cmidrule(lr){6-7}
Row & B-F1 & H-F1 & B-F1 & H-F1 & B-F1 & H-F1\\
\midrule
RallyArb (training-free) & 0.709 & 0.590 & 0.647 & 0.456 & 0.831 & 0.594\\
E2E-Spot, published weights & 0.989 & 1.000 & 0.667 & 0.824 & 0.130 & 0.154\\
T-DEED, published weights & 0.978 & 0.957 & 0.623 & 0.817 & 0.000 & 0.000\\
Gemini 3.8 Flash, zero-shot & 0.415 & 0.667 & 0.427 & 0.532 & 0.525 & 0.475\\
Qwen3-VL-235B, zero-shot & 0.221 & 0.228 & 0.100 & 0.286 & 0.142 & 0.232\\
\midrule
E2E-Spot, fine-tuned LOCO & 0.957 & 0.977 & 0.794 & 0.921 & 0.603 & 0.510\\
\bottomrule
\end{tabular}
\caption{Bounce and hit F1 by camera type, development and test clips of one type pooled, rally-only
scoring. The five rows above the line were not trained on the corpus, but RallyArb's constants were
refined on the development clips, which are two of the three phone and court-level clips. Table~\ref{tab:info}
gives each row's information regime, and the test rows of Table~\ref{tab:clips} give the held-out
reading. Generated by \texttt{make\_figures.py} from \texttt{data/results\_external\_baselines.json};
per-clip values are in Table~\ref{tab:clips}.}
\label{tab:camera}
\end{table}

\begin{table}[!htbp]
\centering\small\setlength{\tabcolsep}{4pt}
\begin{tabular}{lccccccrr}
\toprule
& \multicolumn{2}{c}{Dev (pooled)} & \multicolumn{4}{c}{Test (pooled, one execution)} & \multicolumn{2}{c}{Test errors}\\
\cmidrule(lr){2-3}\cmidrule(lr){4-7}\cmidrule(lr){8-9}
Row & B-F1 & H-F1 & B-F1 & 95\% CI & H-F1 & 95\% CI & Spur. & Conf.\\
\midrule
RallyArb (training-free) & 0.852 & 0.646 & 0.621 & [0.49, 0.84] & 0.459 & [0.40, 0.53] & 26 & 8\\
E2E-Spot, published weights & 0.557 & 0.522 & 0.776 & [0.44, 0.99] & 0.905 & [0.68, 1.00] & 0 & 0\\
T-DEED, published weights & 0.494 & 0.507 & 0.752 & [0.36, 0.97] & 0.859 & [0.55, 0.96] & 6 & 0\\
Gemini 3.8 Flash, zero-shot & 0.519 & 0.523 & 0.406 & [0.37, 0.48] & 0.593 & [0.49, 0.69] & 59 & 9\\
Qwen3-VL-235B, zero-shot & 0.135 & 0.217 & 0.190 & [0.13, 0.22] & 0.257 & [0.23, 0.28] & 256 & 28\\
\midrule
E2E-Spot, fine-tuned LOCO & 0.753 & 0.684 & 0.846 & [0.71, 0.96] & 0.941 & [0.85, 0.99] & 7 & 0\\
\bottomrule
\end{tabular}
\caption{Pooled micro-averaged F1 on each split, with clip-level bootstrap 95\% intervals on test
(10\,000 iterations, seed 20260901, four clips). Spur.\ counts unmatched predictions and Conf.\ counts
predictions matched to an event of the other class. Every test value is a single execution. Generated
by \texttt{make\_figures.py} from the evaluator reports in \texttt{data/}.}
\label{tab:pooled}
\end{table}

\begin{figure}[!htbp]
\centering
\includegraphics[width=\linewidth]{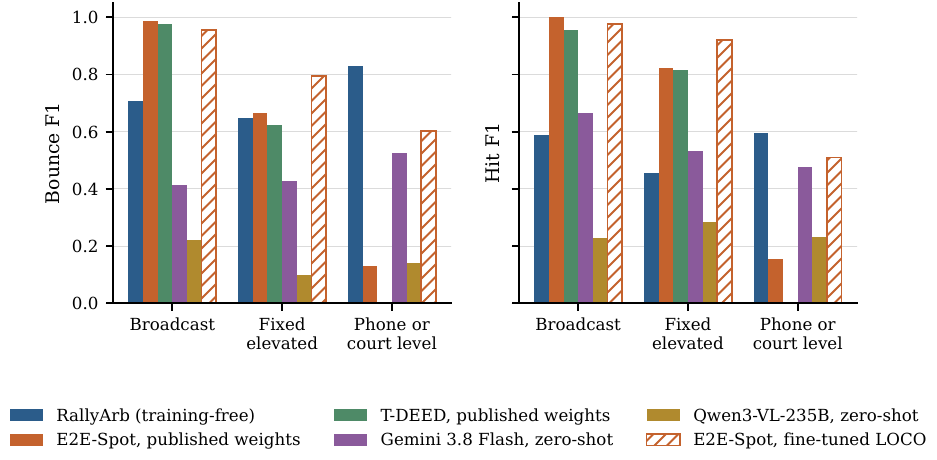}
\caption{Bounce F1 (left) and hit F1 (right) by camera type for the five rows that learned nothing from
the corpus and, hatched, for E2E-Spot fine-tuned leave-one-clip-out on the released events.
Development and test clips of one camera type are pooled. Generated by \texttt{make\_figures.py} from
\texttt{data/results\_external\_baselines.json}.}
\label{fig:barres}
\end{figure}

\paragraph{Result 1 (supervised spotters are precise on broadcast and nearly silent on phone and court-level footage).}
With their published broadcast weights, the supervised spotters score near-perfectly on broadcast clips
and almost nothing on phone and court-level footage. E2E-Spot reaches a pooled bounce and hit F1 of 0.989 and
1.000 on the three broadcast clips, 0.667 and 0.824 on the two fixed elevated cameras, and 0.130 and
0.154 on the three phone and court-level clips. It makes no confusion and no spurious prediction on either split,
so its errors are all misses. T-DEED follows the same profile and scores zero on all three phone and court-level clips. The clay-pro clip, a 2025 professional clay-court broadcast on a surface absent from the
spotters' US Open and Wimbledon training matches, is scored 1.000 and 1.000 by E2E-Spot.

\paragraph{Result 2 (vision-language models detect events on every viewpoint, off the contact frame).}
Gemini 3.8 Flash scores a bounce F1 between 0.415 and 0.525 on every camera type, with 59 spurious
predictions and 9 confusions on the four test clips. On test hits it is ahead of RallyArb, with 0.593
against 0.459, mostly on the two broadcast clips. Qwen3-VL emits 256 spurious predictions on test,
about two per real event, for a pooled test F1 of 0.190 and 0.257. The Gemini test pass cost 2.33~USD.
At a tolerance of 0.04~s the bounce F1 of Gemini falls from 0.406 to 0.195 (Result~6). Its sampling
interval of about 83~ms bounds how close it can place an event, so part of this timing error belongs
to the protocol and not to the model.

\paragraph{Result 3 (the training-free row keeps phone and court-level bounces, on one held-out clip).} On each of
the three phone and court-level clips, RallyArb has a higher bounce F1 than every other row: 0.828 on felix and
hard-pro and 0.842 on grass-am, against 0.000 to 0.235 for the published spotters and 0.41 to 0.65 for
Gemini. Two of these clips are development clips on which its constants were refined. The one held-out clip of this group, grass-am, has 11 rally bounces and 8 rally hits. RallyArb scores 0.842 and
0.556 there. E2E-Spot scores 0.167 and 0.400 and Gemini 0.500 and 0.600, and the fine-tuned spotter
finds 7 of the 11 bounces against 8 for RallyArb. The bounce result therefore holds on the held-out
clip, and the hit result does not, since Gemini leads on grass-am hits. On broadcast clips RallyArb is
less precise than the spotters, with 0.709 and 0.590 and, on the two broadcast test clips, 14 spurious
predictions and 3 confusions.

\paragraph{Result 4 (the pooled ranking follows the composition of the split).} The pooled ranking
inverts between the splits. On the development split, which holds three non-broadcast clips, RallyArb
leads with 0.852 and 0.646 against 0.557 and 0.522 for E2E-Spot. On the test split, where two
broadcast clips carry 72 of 140 events, E2E-Spot leads with 0.776 and 0.905 against 0.621 and 0.459.
The per-camera reading of Table~\ref{tab:camera} does not depend on this composition.

\paragraph{Result 5 (fine-tuning reaches the phone and court-level clips except the grazing one).} Fine-tuned
leave-one-clip-out, E2E-Spot reaches a pooled test F1 of 0.846 and 0.941 with no confusion. These are
cross-validation scores, since each test fold trains on the other test clips. The fine-tuned model
overtakes RallyArb on the fixed elevated cameras (0.794 and 0.921 against 0.647 and 0.456). On hard-pro
it scores 0.800 and 0.714, against 0.235 and 0.000 for the published E2E-Spot, and on grass-am 0.778
and 0.714 against 0.167 and 0.400. On the grazing angle of felix it scores 0.200 and 0.261, against
0.828 and 0.500 for RallyArb.

\begin{figure}[!htbp]
\centering
\includegraphics[width=\linewidth]{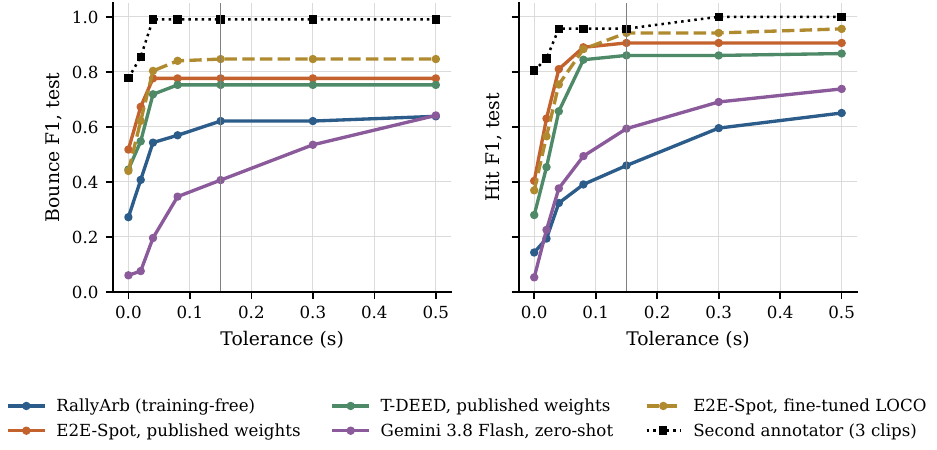}
\caption{Pooled test F1 against the matching tolerance, applied per clip as $\mathrm{round}(t\,\nu)$
frames, rally-only view. The vertical line marks the paper tolerance of 0.15~s. The dotted black curve
scores the second annotator against the first on the three double-annotated clips (one development
clip and two test clips; the system curves cover the four test clips), with the second annotator's own
tagged events removed before matching; an event tagged only by the second annotator counts as a miss,
which gives 0.990 bounce F1 where the symmetric agreement of Section~\ref{sec:banc} gives 1.000. The
spotters run on every second frame of 50 and 60 frames per second clips, which bounds their exact-frame
score. Generated by \texttt{make\_figures.py} from \texttt{data/tolerance\_sweep.json}.}
\label{fig:tol}
\end{figure}

\paragraph{Result 6 (the spotters' matches lie closest to the annotated frame).}
Figure~\ref{fig:tol} scores every row at tolerances from 0 to 0.5~s. The second annotator's agreement
with the first is 0.777 and 0.804 at the exact frame and 0.990 and 0.957 at 0.04~s. The published
E2E-Spot reaches its 0.15~s bounce score at 0.04~s (0.776), while its hit F1 is 0.810 there against
0.905 at 0.15~s; the second annotator's curve covers three clips, one of them from the development
split, and the system curves cover the four test clips. RallyArb's hit F1 keeps rising past the paper
tolerance, from 0.459 at 0.15~s to 0.595 at 0.30~s, and Gemini's bounce F1 rises from 0.406 to 0.534.
On same-class test matches within 0.5~s, RallyArb places hits 0.105~s from the contact frame on
average, Gemini places bounces and hits 0.145 and 0.093~s from it and Qwen 0.130 and 0.116~s, against
0.009 and 0.019~s for E2E-Spot. Tight tolerances do not remove the result on this group. At 0.04~s, the pooled bounce F1 of this group is 0.769 for RallyArb, 0.554 for the fine-tuned spotter and 0.217 for
Gemini, and on the held-out grass-am alone it is 0.700, 0.700 and 0.200. RallyArb's hits are anchored
on wrist-speed peaks, which are not the contact frame, and part of its hit deficit at 0.15~s is a
timing error rather than a missed event.

\section{Threats to validity}
\paragraph{Annotation.} One person annotated all eight clips and also designed the arbitration layer.
The version-2 tag pass on the test split was blind to predictions, but the policy change it applied
came from an audit of RallyArb's own spurious predictions on the development clips. The second
annotation measures the reliability of the labels on three clips, at 1.000 and 0.957 in the rally-only
view with frame offsets of at most 2, which is small against tolerances of 4 to 9 frames. The second
annotator is a colleague at the authors' company. The two events on which the annotators disagree are
development events that the first annotator retimed after reading spurious predictions, and one of
the retimed frames coincides with a RallyArb prediction. Moving both back to the second annotator's
frames lowers the development hit F1 of RallyArb from 0.646 to 0.626 and that of E2E-Spot from 0.522
to 0.493. No test event was retimed, and five of the eight clips have a single annotation.

\paragraph{Sample size.} Eight clips give four per split and two or three per camera type. The
bootstrap intervals are wide; the test bounce interval of E2E-Spot spans 0.44 to 0.99. Test intervals
are up to 0.61 wide, and with three or four clips per group no two-sided sign test over clips can reach
$p<0.125$. The direction of the viewpoint effect is readable because it holds on each of the three
phone and court-level clips separately, with bounce F1 gaps of 0.59 to 0.83 against E2E-Spot. Smaller
differences, including RallyArb against the fine-tuned spotter on grass-am, are within one or two
events. The claim for this group for RallyArb rests on one held-out clip (Result~3).

\paragraph{Training-set overlap.} The published spotters were trained on 28 US Open and Wimbledon
broadcasts collected from YouTube (Hong et al., 2022). Demo3 and grass-pro are professional broadcasts
of the same kind as these matches, so the spotters' broadcast scores may benefit from in-distribution
training data. Clay-pro, a 2025 professional clay-court broadcast, lies outside that distribution, and
the spotters score it as high as the other two broadcast clips.

\paragraph{Information regimes.} The rows do not see the same information (Table~\ref{tab:info}). Each
test fold of the fine-tuned spotter trains on the labels of the three other test clips. The
fine-tuning results hold for the smaller RegNetY-002 model and these settings only.

\paragraph{External systems and metrics.} The supervised spotters were run with published weights and
the published input recipe, which reads every second frame of our 50 and 60 frames per second clips and
a centre crop of 224 pixels; the ball lies outside this crop at 15 of the 32 tracked rally events of
felix, 6 of 21 on hard-pro and 3 of 18 on grass-am, so part of the spotters' silence on phone and court-level clips may come from the crop. Precise-spotting papers report average precision at tolerances of 0 to
2 frames. The frozen predictions are thresholded, and RallyArb and the vision-language models emit no
score, so we report F1 at several tolerances (Result~6) and no average precision. The vision-language
models are dated API versions of 2026-10-15 to 2026-10-17, and their prompt, sampling rate and the
choice of Gemini among six models were set on the development split. These are one reasonable
configuration each, and a better prompt or denser sampling could raise the vision-language rows.

\paragraph{Upstream components.} RallyArb's scores bound the whole pipeline, including the ball tracker
and the pose estimator. Its rows therefore measure the layer together with its inputs and would change
with a better tracker.

\section{Reproducibility}
\paragraph{Release.} The benchmark repository\footnote{\url{https://github.com/David-dsv/tennis-bounce-hit-benchmark};
data CC~BY~4.0, code MIT.} holds the eight ground-truth sidecars, the annotation policy, the matcher
and the multi-clip evaluator, the replay fixtures, the frozen predictions of every row with one
manifest per row, the evaluator reports each manifest should reproduce, the double-annotation protocol
and its results. The scripts that build the tables and figures of this paper, the tolerance sweep and
the timing analysis are released with the paper sources as ancillary files.

\paragraph{Regimes.} Every number in the tables and figures is recomputed from released files by the
standard-library evaluator. Scoring the frozen predictions is exact: on 2026-10-26 the released
evaluator reproduced every pooled number of Tables~\ref{tab:camera} and~\ref{tab:pooled} to the last
digit.

\paragraph{Commands.} The evaluator needs Python 3.9 or later and no other package. From the
repository root,
\begin{verbatim}
python eval/eval_multiclip.py --manifest eval/manifest.json --split test \
    --exclude-tags serve,toss,far,dead,offscreen,replay
\end{verbatim}
scores RallyArb, and the same command with \texttt{eval/baselines/<row>/manifest.json} scores each
other row. The tables and figures are rebuilt by \texttt{python3 make\_figures.py}.

\paragraph{Videos and code.} The source videos are not redistributed, and the replay fixtures allow any
method that works downstream of tracking to be scored without them. Appendix~\ref{app:arb} specifies
the arbitration layer, and its code is available from the authors on request.

\section{Conclusion}
On a small benchmark that mixes camera viewpoints, the ranking of bounce and hit detectors follows the
camera. Supervised spotters with broadcast weights are precise on broadcast footage and nearly silent on phone and court-level footage. Vision-language models detect events on every viewpoint and place them about 0.1~s
from the contact. A training-free arbitration layer keeps a bounce F1 above 0.8 on all three
phone and court-level clips, the held-out one included, but places its hits too loosely. Fine-tuned on the
released events, the supervised spotter learns every viewpoint of the corpus except the grazing one.
The benchmark measures whether a detector holds across viewpoints, and extending it to more
court-level and phone viewpoints is the next step.

\section*{Acknowledgements}
We thank Paul Lemaistre for the second annotation.

\section*{References}
\small
\begin{itemize}\setlength{\itemsep}{2pt}\renewcommand{\labelitemi}{}\setlength{\leftmargin}{0pt}
\item S. Agrawal, R. Sundararajan and V. Sagar. Accurate tennis court line detection on amateur recorded matches, 2024. arXiv:2404.06977.
\item J. Bian, X. Li, T. Wang, Q. Wang, J. Huang et al. P2ANet: A dataset and benchmark for dense action detection from table tennis match broadcasting videos, 2022. arXiv:2207.12730.
\item Y.-J. Chen and Y.-S. Wang. TrackNetV3: Enhancing shuttlecock tracking with augmentations and trajectory rectification. In \emph{Proc. ACM MMAsia}, 2023. doi:10.1145/3595916.3626370.
\item A. Cioppa, S. Giancola, V. Somers, F. Magera, X. Zhou et al. SoccerNet 2023 challenges results, 2023. arXiv:2309.06006.
\item A. Cioppa, S. Giancola, V. Somers, V. Joos, F. Magera et al. SoccerNet 2024 challenges results, 2024. arXiv:2409.10587.
\item A. Deliège, A. Cioppa, S. Giancola, M. J. Seikavandi, J. V. Dueholm, K. Nasrollahi, B. Ghanem, T. B. Moeslund and M. Van Droogenbroeck. SoccerNet-v2: A dataset and benchmarks for holistic understanding of broadcast soccer videos. In \emph{Proc. CVPRW}, 2021. arXiv:2011.13367.
\item D. Deng, J. Wu, J. Wang, Y. Wu, X. Xie, Z. Zhou, H. Zhang, X. Zhang and Y. Wu. EventAnchor: Reducing human interactions in event annotation of racket sports videos. In \emph{Proc. CHI}, 2021.
\item L. Dong, Y. Yang, H. Wu, W. Wang, Y. Hou, Z. Zhong and X. Sun. RacketVision: A multiple racket sports benchmark for unified ball and racket analysis. In \emph{Proc. AAAI}, 2026. arXiv:2511.17045.
\item D. Farin, S. Krabbe, P. H. N. de With and W. Effelsberg. Robust camera calibration for sport videos using court models. In \emph{Storage and Retrieval Methods and Applications for Multimedia}, volume 5307 of \emph{Proc. SPIE}, pages 80--91, 2004.
\item H. Faulkner and A. Dick. TenniSet: A dataset for dense fine-grained event recognition, localisation and description. In \emph{Proc. DICTA}, 2017. doi:10.1109/DICTA.2017.8227494.
\item S. Giancola, A. Cioppa, M. Gutiérrez-Pérez, J. Held, C. Hinojosa et al. SoccerNet 2025 challenges results, 2025. arXiv:2508.19182.
\item J. Hong, H. Zhang, M. Gharbi, M. Fisher and K. Fatahalian. Spotting temporally precise, fine-grained events in video. In \emph{Proc. ECCV}, 2022. arXiv:2207.10213.
\item Y.-C. Huang, I.-N. Liao, C.-H. Chen, T.-U. \.Ik and W.-C. Peng. TrackNet: A deep learning network for tracking high-speed and tiny objects in sports applications. In \emph{Proc. IEEE AVSS}, 2019.
\item M. Jung, J. Xiao, B.-T. Zhang and A. Yao. On the consistency of video large language models in temporal comprehension. In \emph{Proc. CVPR}, 2025. arXiv:2411.12951.
\item R. E. Kalman. A new approach to linear filtering and prediction problems. \emph{J. Basic Engineering}, 82(1):35--45, 1960.
\item P. Liu and J.-H. Wang. MonoTrack: Shuttle trajectory reconstruction from monocular badminton video. In \emph{Proc. CVPRW}, pages 3512--3521, 2022.
\item Z. Liu, K. Jiang, M. Ma, Z. Hou, Y. Lin and J. S. Dong. F\textsuperscript{3}Set: Towards analyzing fast, frequent, and fine-grained events from videos. In \emph{Proc. ICLR}, 2025. arXiv:2504.08222.
\item Z. Liu, K. Jiang, M. Ma, Z. Hou, Y. Lin and J. S. Dong. Few-shot precise event spotting via unified multi-entity graph and distillation. In \emph{Proc. AAAI}, 2026. arXiv:2511.14186.
\item I. Loshchilov and F. Hutter. Decoupled weight decay regularization. In \emph{Proc. ICLR}, 2019.
\item N. Owens, C. Harris and C. Stennett. Hawk-eye tennis system. In \emph{Proc. VIE}, pages 182--185, 2003.
\item I. Radosavovic, R. P. Kosaraju, R. Girshick, K. He and P. Dollár. Designing network design spaces. In \emph{Proc. CVPR}, 2020.
\item A. Raj, L. Wang and T. Gedeon. TrackNetV4: Enhancing fast sports object tracking with motion attention maps, 2024. arXiv:2409.14543.
\item P. K. Sen. Estimates of the regression coefficient based on Kendall's tau. \emph{J. American Statistical Association}, 63(324):1379--1389, 1968.
\item N.-E. Sun, Y.-C. Lin, S.-P. Chuang, T.-H. Hsu, D.-R. Yu, H.-Y. Chung and T.-U. \.Ik. TrackNetV2: Efficient shuttlecock tracking network. In \emph{Proc. ICPAI}, pages 86--91, 2020.
\item S. Tarashima, M. A. Haq, Y. Wang and N. Tagawa. Widely applicable strong baseline for sports ball detection and tracking. In \emph{Proc. BMVC}, 2023.
\item H. Theil. A rank-invariant method of linear and polynomial regression analysis. \emph{Proc. Koninklijke Nederlandse Akademie van Wetenschappen}, 53:386--392, 1950.
\item A. Viterbi. Error bounds for convolutional codes and an asymptotically optimum decoding algorithm. \emph{IEEE Trans. Information Theory}, 13(2):260--269, 1967.
\item R. Voeikov, N. Falaleev and R. Baikulov. TTNet: Real-time temporal and spatial video analysis of table tennis. In \emph{Proc. CVPRW}, 2020.
\item W.-Y. Wang, Y.-C. Huang, T.-U. Ik and W.-C. Peng. ShuttleSet: A human-annotated stroke-level singles dataset for badminton tactical analysis. In \emph{Proc. KDD}, 2023. arXiv:2306.04948.
\item A. Xarles, S. Escalera, T. B. Moeslund and A. Clapés. T-DEED: Temporal-discriminability enhancer encoder-decoder for precise event spotting in sports videos. In \emph{Proc. CVPRW}, pages 3410--3419, 2024. arXiv:2404.05392.
\item H. Xu, A. Agrahari Baniya, S. Well, M. R. Bouadjenek, R. Dazeley and S. Aryal. Deep learning for sports video event detection: Tasks, datasets, methods, and challenges, 2025. arXiv:2505.03991.
\item H. Zhang, C. Sciutto, M. Agrawala and K. Fatahalian. Vid2Player: Controllable video sprites that behave and appear like professional tennis players. \emph{ACM Trans. Graphics}, 40(3), 2021.
\end{itemize}
\normalsize

\appendix
\section{Per-clip scores}
\begin{table}[h]
\centering\small\setlength{\tabcolsep}{4pt}
\begin{tabular}{llccccccc}
\toprule
Clip & Split & RallyArb & E2E-Spot & T-DEED & Gemini & Qwen & E2E LOCO\\
\midrule
demo3 & dev & 1.00/1.00 & 1.00/1.00 & 1.00/0.94 & 0.43/0.59 & 0.24/0.12 & 1.00/1.00\\
felix & dev & 0.83/0.50 & 0.00/0.10 & 0.00/0.00 & 0.41/0.29 & 0.00/0.22 & 0.20/0.26\\
clay-am & dev & 0.81/0.54 & 0.81/0.75 & 0.74/0.77 & 0.53/0.67 & 0.09/0.32 & 0.93/0.85\\
hard-pro & dev & 0.83/0.78 & 0.24/0.00 & 0.00/0.00 & 0.65/0.62 & 0.18/0.19 & 0.80/0.71\\
clay-pro & test & 0.49/0.49 & 1.00/1.00 & 1.00/0.98 & 0.39/0.67 & 0.23/0.27 & 0.98/1.00\\
clay-am2 & test & 0.50/0.38 & 0.55/0.86 & 0.53/0.84 & 0.36/0.45 & 0.11/0.26 & 0.68/0.96\\
grass-pro & test & 0.83/0.48 & 0.95/1.00 & 0.91/0.93 & 0.45/0.71 & 0.20/0.22 & 0.87/0.92\\
grass-am & test & 0.84/0.56 & 0.17/0.40 & 0.00/0.00 & 0.50/0.60 & 0.20/0.29 & 0.78/0.71\\
\bottomrule
\end{tabular}
\caption{Bounce F1 / hit F1 per clip and row, rally-only scoring. RallyArb's constants were fixed on
demo3 and refined on the four development clips. Every test value is a single execution. Generated by
\texttt{make\_figures.py}.}
\label{tab:clips}
\end{table}

\section{Additional released rows}
The repository holds three further rows, scored by the same evaluator, from which the paper draws no
conclusion. A camera-type rule that routes broadcast and fixed elevated clips to E2E-Spot and
phone and court-level clips to RallyArb scores 0.846 and 0.896 on test; it was written after both components had
been scored on test. Adding the zero-shot spotter's bounce frames to RallyArb's candidate pool, in a
configuration selected on the development split, scores 0.704 and 0.459 on test against 0.621 and
0.459 without them. The leave-one-clip-out fine-tuning started from ImageNet weights, with the same
settings and a new three-way head, emits no prediction above the threshold on any fold.

\section{The arbitration layer}\label{app:arb}
This appendix describes the configuration scored on test.

\paragraph{Inputs.} The layer reads a ball track with a per-frame validity mask, a spline-completed copy
used for lookup only, and two identity-locked players with boxes and 17 keypoints. All directional
evidence is read on real detections, never on the spline, and a measurement that cannot be read
abstains.

\paragraph{Candidates.} The bounce detector fits iteratively reweighted lines on each side of a
candidate apex, with hard rejection beyond $2.5\sigma$, and keeps a candidate that passes slope,
residual, slope-restitution ($\le 2.5$) and energy-ratio ($\le 1.8$) tests. The hit detector keeps
wrist-speed peaks per side of the net that pass a full-width-at-half-maximum gate and a per-side
amplitude floor scaled by the median player-box height. Three trajectory generators add local extrema
of the image height (prominence $0.010\,H$, separation $0.15\,\nu$), velocity turns of at least
$70^\circ$, and near-court knees where a steep descent settles to near-zero vertical velocity.
Candidates within $0.18\,\nu$ frames of a cluster's start join that cluster.

\paragraph{Evidence.} Over half-windows of $\max(2,\mathrm{round}(0.10\,\nu))$ frames, Theil--Sen slopes
give the horizontal and vertical velocity on each side of the cluster. Floor evidence is a vertical
flip from descending to ascending with the horizontal direction preserved. The contact distance is the
distance from the tracked ball to the nearest confident wrist or box centre, divided by that player's
box height, with bands of 0.5 (at the wrist) and 1.2 (clear separation). Measured at the ball, real
contacts on the development rally lie at or below 0.51 and phantom swing anchors at or above 0.90.
Measuring at the wrist position reported by the hit detector would be circular, since every wrist-speed
peak is near a wrist by construction.

\paragraph{Label legality.} A cluster is hit-like when it carries a corroborated swing, the ball at a
wrist with an incoming ball, or a confident trajectory redirect, and no floor evidence. A hit-like
cluster cannot be labeled a bounce, and a cluster with clear floor evidence and no swing or redirect
cannot be labeled a hit. These rules remove label hypotheses before scoring, so no score can restore
them. Clusters with floor evidence become mandatory bounce anchors, clusters with a corroborated swing
at the wrist become hit anchors, and both receive the same bonus in the decoder.

\paragraph{Decoding.} The clip's beat is the median of anchor gaps, each divided by its rounded beat
count from a seed of $0.55\,\nu$. A dynamic program decides, for each cluster, to skip it or to emit a
label. Across a gap $g$ it considers both bracketing beat counts $k$, requires the label to flip
exactly when $k$ is odd, forbids an immediate repeat of one class, and charges
$0.5\,|g-k\hat\Delta|/\hat\Delta+(k-1)$. Mandatory anchors cannot be skipped. The beam keeps 256
states.

\end{document}